\documentclass[letterpaper, 10 pt, conference]{ieeeconf}

\IEEEoverridecommandlockouts
\title{\LARGE \bf
ELMP: Efficient Learning for Motion Planning \\ via Analytical Policy Gradients
}

\author{Yixiao Li$^1$, Tifanny Portela$^1$, Jordis Herrmann$^{2,\dagger}$, René Zurbrügg$^1$, Marco Hutter$^1$%
\thanks{This work was primarily supported by the ETH AI Center. $^1$Robotic Systems Lab, ETH Z\"{u}rich; $^2$ABB. $^\dagger$Jordis Herrmann is currently with ANYbotics.}%
\thanks{Email: {\tt\small \{yixili,tportela\}@ethz.ch}}%
}
\usepackage{booktabs}
\usepackage{multirow}
\usepackage[table]{xcolor}
\usepackage{graphicx}
\usepackage{mathtools}
\usepackage{algorithm}
\usepackage{algorithmicx}
\usepackage{algpseudocode}
\usepackage{amsmath}
\usepackage{lettrine}
\usepackage{cite}
\usepackage{setspace}
\usepackage{amssymb}
\usepackage{threeparttable}
\usepackage{caption}
\begin{document}

\maketitle
\thispagestyle{empty}
\pagestyle{empty}

\begin{abstract}
Neural Motion Planners (NMPs) enable fast reactive motion generation, but adapting them to new environments typically requires recollecting large expert datasets, which is computationally prohibitive.
We propose \textbf{ELMP}, a framework for data-efficient adaptation via self-supervised fine-tuning. Rather than generating additional expert trajectories with expensive global planners, ELMP directly optimizes the policy through a differentiable kinematic layer using dense collision, target-reaching, and smoothness objectives. This replaces expert data generation with rapid problem sampling, reducing per-sample adaptation cost by roughly two orders of magnitude.
To further support robust generalization across changing kinematic chains, we introduce a mechanism to explicitly encode tool geometry via point clouds.
Benchmarked against classical and neural baselines, ELMP achieves an 84.8\% average success rate with orders-of-magnitude lower cold-start latency than classical methods. In unseen environments, self-supervised fine-tuning improves success rate from 57.3\% (zero-shot) to 89.8\%, removing the data collection bottleneck.
Our approach maintains millisecond-level inference latency and is validated on a physical Franka Emika Panda robot.
\end{abstract}


\section{Introduction}

Generating collision-free motion is fundamental to robotic manipulation but remains challenging in unstructured environments~\cite{fishman2023motion, dalal2024neural, yang2025deep}.
Traditional sampling-based~\cite{lavalle1998rapidly, kavraki1996probabilistic, bohlin2000path, gammell2020batch, strub2020adaptively} and optimization-based~\cite{ratliff2009chomp, schulman2014motion, liu2018convex, sundaralingam2023curobo} methods provide completeness or optimality guarantees but incur high latency from their sequential sense-plan-act pipeline, often precluding real-time replanning.
Local reactive methods such as STORM~\cite{bhardwaj2022storm} and Geometric Fabrics~\cite{van2022geometric} achieve high-frequency control but lack long-horizon foresight, making them vulnerable to local minima in complex geometries.

\begin{figure}[t]
\centering
\includegraphics[width=0.47\textwidth]{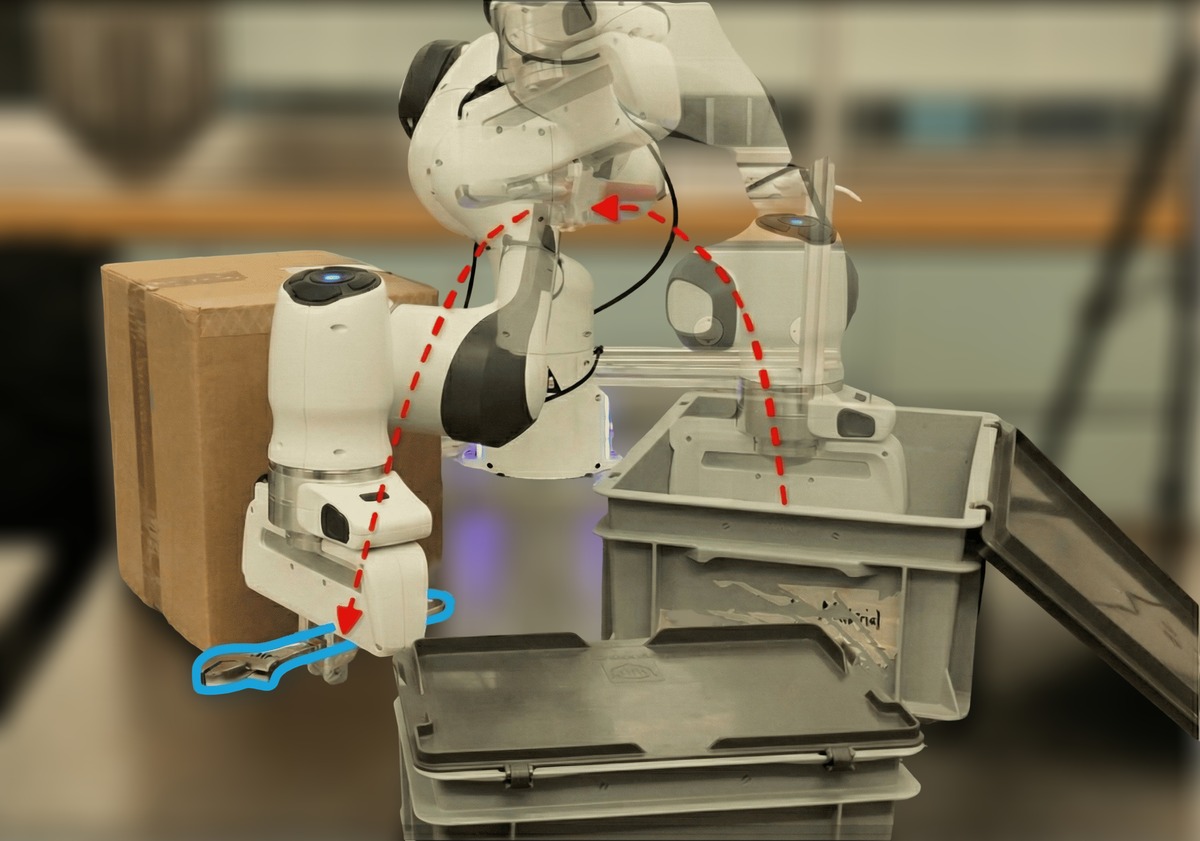}
\caption{\textbf{Tool-Aware Manipulator Motion Planning}: A Franka Emika Panda robot executes a pick-and-place task involving a tool (wrench). Our method ELMP explicitly encodes the variable tool geometry via point clouds to enable collision avoidance for the entire kinematic chain, while leveraging Analytical Policy Gradients to fine-tune the policy for high-precision, collision-free motion.}
\label{fig:snap}
\vspace{-5mm}
\end{figure}

Neural Motion Planners (NMPs)~\cite{fishman2023motion, dalal2024neural, yang2025deep} compress planning into a neural network for fast inference, but standard Behavior Cloning (BC) suffers from covariate shift over long horizons, requiring prohibitive amounts of expert data.
Moreover, most NMPs assume fixed robot geometry, neglecting variable end-effectors or grasped objects.
Although recent methods such as Neural MP~\cite{dalal2024neural} have begun addressing variable embodiments, they lack extensive evaluation of point-cloud representations, and bounding-box abstractions~\cite{lee2025learning} sacrifice geometric fidelity, failing in cluttered spaces.
For tool-aware manipulation (Fig.~\ref{fig:snap}), explicit geometric reasoning about the full kinematic chain is required.

We introduce \textbf{ELMP} (\textbf{E}fficient \textbf{L}earning for \textbf{M}otion \textbf{P}lanning via Analytical Policy Gradients), a two-stage neural planner: BC pre-training followed by self-supervised APG fine-tuning through a differentiable kinematic layer with dense collision, target-reaching, and smoothness objectives. This enables adaptation to novel environments without additional expert demonstrations. We further encode tool geometry as a point cloud, conditioning the policy to generate collision-free trajectories for the entire kinematic chain.
While ELMP builds on established components (PointNet++ encoders, SDF-based objectives), it is the first to combine self-supervised APG fine-tuning with explicit tool-aware conditioning for arm motion planning, enabling data-efficient environment transfer.

In summary, our contributions are as follows:
\begin{itemize}

    \item \textbf{Self-supervised adaptation via APG:} We introduce Analytical Policy Gradient fine-tuning to kinematic neural motion planning, enabling adaptation to new environments without re-collecting additional demonstrations from computationally expensive expert planners.

    \item \textbf{Tool-aware geometric conditioning:} We demonstrate that explicit point-cloud conditioning of the entire kinematic chain, including variable tools, equips the planner to safely plan through cluttered environments, significantly outperforming coarse bounding-box abstractions.

    \item \textbf{Comprehensive empirical validation:} We evaluate on held-out simulation benchmarks and on novel environments outside the pre-training distribution, where self-supervised fine-tuning substantially improves success. We additionally demonstrate feasibility on a real robot.
\end{itemize}

\section{Related Work}

\textbf{Robot Motion Planning:} Traditional collision-free planning relies on sampling-based methods (e.g. RRT$^*$ \cite{lavalle1998rapidly}, AIT$^*$ \cite{strub2020adaptively}) or optimization-based approaches (e.g. CHOMP \cite{ratliff2009chomp}, CuRobo \cite{sundaralingam2023curobo}). Although these provide completeness or smoothness guarantees, they suffer from high computational variance or variable ``cold-start'' latencies that hinder real-time replanning \cite{sundaralingam2023curobo}, and solve each query from scratch without learning a reusable policy. In contrast, local reactive controllers such as STORM \cite{bhardwaj2022storm} and Geometric Fabrics \cite{van2022geometric} achieve high-frequency avoidance but lack long-horizon foresight, leaving them vulnerable to getting trapped in local minima in complex geometries.

\textbf{Neural Motion Planning:} Neural Motion Planners (NMPs) such as Motion Planning Networks~\cite{qureshi2021motion}, M$\pi$Nets \cite{fishman2023motion}, and Neural MP \cite{dalal2024neural} address these latency bottlenecks by mapping sensory input directly to actions. Motion Planning Networks~\cite{qureshi2021motion} pioneered learning-based planners that bridge classical and neural planning via iterative bidirectional planning, but do not address variable tool geometry or self-supervised gradient-based adaptation. However, relying predominantly on Behavior Cloning (BC) makes these policies highly susceptible to covariate shift over long horizons \cite{yang2025deep}. Correcting this often requires computationally expensive oracle planners for fine-tuning. Furthermore, most existing NMPs assume fixed robot geometry, lacking explicit mechanisms like point clouds to handle variable tool shapes zero-shot.

ELMP builds on the M$\pi$Nets architecture but adds APG-based adaptation and tool-aware point-cloud conditioning to address both limitations. Recent methods push the frontier of generalist neural planning at the cost of massive data scale: Neural MP~\cite{dalal2024neural} distills 1--3M expert trajectories and employs test-time optimization (TTO) for out-of-distribution adaptation, incurring multi-second latency; Deep Reactive Policy~\cite{yang2025deep} trains on $\sim$10M trajectories for reactive planning in partially observable scenes but does not support variable tools. ELMP targets a complementary regime---data-efficient adaptation from $\sim$600K trajectories via self-supervised APG, with millisecond-level feed-forward inference and explicit tool-awareness.

\textbf{Differentiable Physics and Policy Optimization:} Differentiable simulation has enabled sample-efficient policy learning via analytic gradients \cite{freeman2021brax, newton, pan2025learning}. Although analytical policy gradients (APG) successfully optimize low-level dynamic controllers by backpropagating performance errors over time \cite{Wiedemann2023Training}, their application to kinematic motion planning remains underexplored. ELMP bridges this gap by implementing a differentiable kinematic layer. By analytically computing Signed Distance Field (SDF) gradients, our framework fine-tunes policies directly against dense planning objectives---bypassing the need for ground-truth expert demonstrations. A related line of work learns neural SDF representations for navigation: Bukhari et al.~\cite{bukhari2025differentiable} propose differentiable composite neural SDFs for dynamic indoor scenes, which is highly relevant to our differentiable collision objective. Our current implementation uses analytical primitive SDFs for efficiency and exact gradients; neural/composite SDFs represent a promising route to handle non-primitive obstacles (see Sec.~\ref{sec:conclusion}).

\section{Method}
\begin{figure*}[tb]
\centering
{\includegraphics[width=1\textwidth]{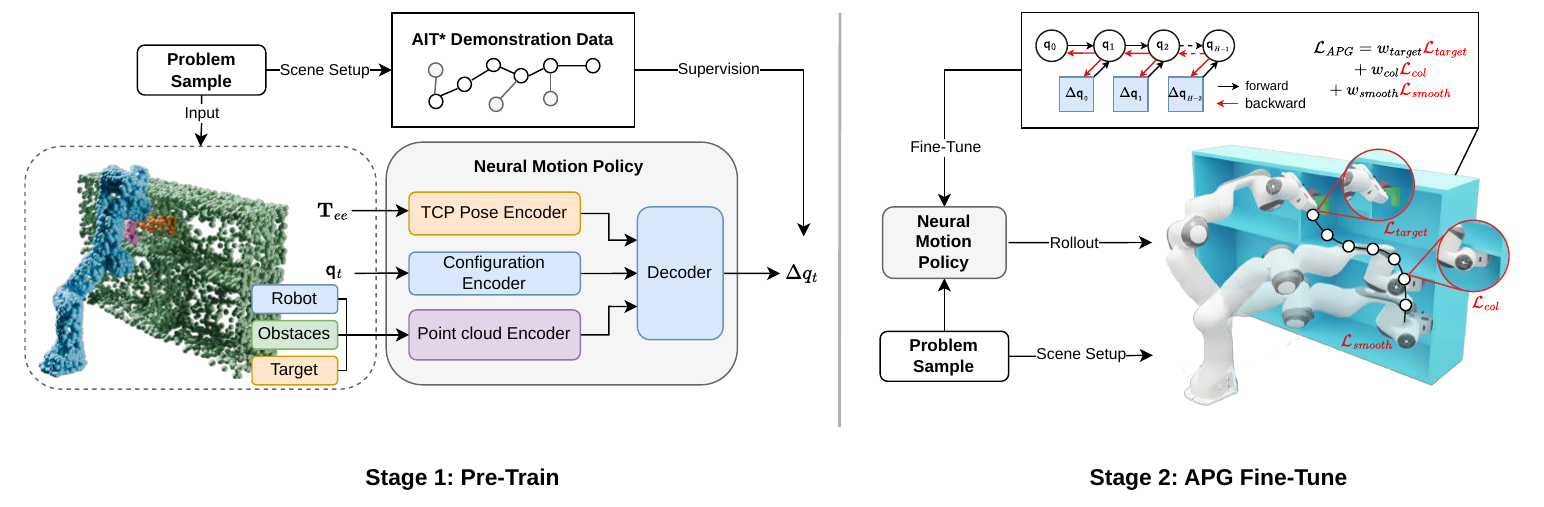}}
\caption{\textbf{Method Overview:} We present ELMP, a framework consisting of two main stages. \textbf{Stage 1 Pre-Train:} A neural motion policy encodes semantic point clouds (robot, obstacles, target), proprioceptive configurations, and TCP poses to predict joint increments, supervised by AIT* demonstration data. \textbf{Stage 2 APG Fine-Tune:} A self-supervised optimization loop where the pre-trained policy is unrolled over horizon $H$ via a differentiable kinematic rollout. Gradients from a differentiable loss (comprising target, collision, and smoothness objectives) are backpropagated through time to fine-tune the policy.}
\label{fig:method_overview}
\vspace{-4mm}
\end{figure*}

In this section, we present the ELMP framework. We first define the formulation of the motion planning problem. Then, we describe the neural policy architecture and detail the self-supervised Analytical Policy Gradient fine-tuning process used to adapt the policy to new environments. Finally, we introduce our differentiable scene model and procedural data generation pipeline.

\subsection{Problem Formulation}

We formulate the motion planning problem as a sequential decision-making task in state space. At each time step $t$, the policy observes the current joint configuration $\mathbf{q}_t \in \mathbb{R}^7$, the target pose of the end-effector $\mathbf{T}_{ee}$, and the observation of point clouds $\mathbf{P}_t \in \mathbb{R}^{N \times 4}$. The policy outputs an increase in the joint position $\Delta \mathbf{q}_t \in \mathbb{R}^7$, resulting in the next state $\mathbf{q}_{t+1} = \mathbf{q}_t + \Delta \mathbf{q}_t$. The objective is to auto-regressively generate a trajectory $\tau = \{\mathbf{q}_0, \dots, \mathbf{q}_H\} \in \mathbb{R}^{H \times 7}$ of the horizon $H$ that reaches $\mathbf{T}_{ee}$ while ensuring that the entire kinematic chain, including the attached tool, avoids collisions with the environment.

\subsection{Policy Architecture and Pre-training}

We adopt a neural policy architecture inspired by M$\pi$Nets \cite{fishman2023motion}, designed to process input of high-dimensional point cloud alongside proprioceptive states.

\textbf{Input Representation:} The network input consists of visual and proprioceptive components, following an architecture similar to prior neural planners~\cite{fishman2023motion, dalal2024neural}. The visual input is a segmented point cloud $\mathbf{P}_t \in \mathbb{R}^{N \times 4}$ composed of three distinct sets: $\mathbf{P}_{robot}$, $\mathbf{P}_{scene}$, and $\mathbf{P}_{target}$. To explicitly encode these semantics, each point in $\mathbf{P}_t$ is augmented with an additional feature channel indicating its category.

$\mathbf{P}_{robot} \in \mathbb{R}^{N_r \times 4}$ represents the current configuration of the robot, including the geometry of the attached tool. $\mathbf{P}_{scene} \in \mathbb{R}^{N_s \times 4}$ represents point clouds from the obstacles. $\mathbf{P}_{target} \in \mathbb{R}^{N_t \times 4}$ represents the ``virtual'' robot end-effector and tool at the target pose $\mathbf{T}_{ee}$. In particular, $\mathbf{P}_{robot}$ and $\mathbf{P}_{target}$ are not derived from external perception but are directly sampled from the robot and tool's mesh models using the current joint configuration (proprioception). This decoupled design intentionally bridges the sim-to-real gap. While $\mathbf{P}_{scene}$ is generated from simulated primitives during training, it is populated directly by an external depth camera during physical deployment. Combining these noisy external sensor data with precise, internally generated $\mathbf{P}_{robot}$ creates a robust hybrid perception scheme, ensuring that the policy maintains an accurate understanding of its own kinematics and tool geometry even under severe visual occlusion.

\textbf{Architecture and Encoders:} As shown in Stage 1 of Fig.~\ref{fig:method_overview}, the aggregated semantic point cloud $\mathbf{P}_t$ is processed by a PointNet++ \cite{qi2017pointnet++} encoder to extract geometric features. Concurrently, proprioceptive and pose states are processed via Multi-Layer Perceptrons (MLP). These features are concatenated and passed to an MLP decoder to predict the joint increment $\Delta\mathbf{q}_t$. This explicit decoupling of pose and geometry constitutes our primary architectural deviation from M$\pi$Nets \cite{fishman2023motion}.

\textbf{Pre-training Loss:} We initialize the policy $\pi_\theta$ with Behavior Cloning (BC) using a combined loss inspired by \cite{fishman2023motion}:

\begin{equation}
\label{eq:bc_loss}
\begin{aligned}
\mathcal{L}_{BC} ={}& \lambda_{\text{action}} \left\lVert \Delta \mathbf{q}_{\text{pred}} - \Delta \mathbf{q}_{\text{gt}} \right\rVert^2
\\
&+ \lambda_{\text{fk}} \sum_{i=1}^{M}  \left\lVert \mathrm{FK}_i(\mathbf{q}_{\text{pred}}) - \mathrm{FK}_i(\mathbf{q}_{\text{gt}}) \right\rVert_2 .
\end{aligned}
\end{equation}
Here, $\Delta \mathbf{q}_{\text{pred}}$ and $\Delta \mathbf{q}_{\text{gt}}$ denote the predicted and ground-truth joint increments, respectively, and $\lambda_{\text{action}}$, $\lambda_{\text{fk}}$ are scalar loss weights. $\mathrm{FK}_i(\cdot)$ represents the forward kinematics mapping from the joint configuration to the $i$-th center of the $M$ robot and tool spheres. This objective jointly penalizes discrepancies in predicted joint actions and inconsistencies in the corresponding task-space geometry.
\begin{figure*}[t!]
\centering
   \makebox[\textwidth][c]{\includegraphics[width=0.88\textwidth]{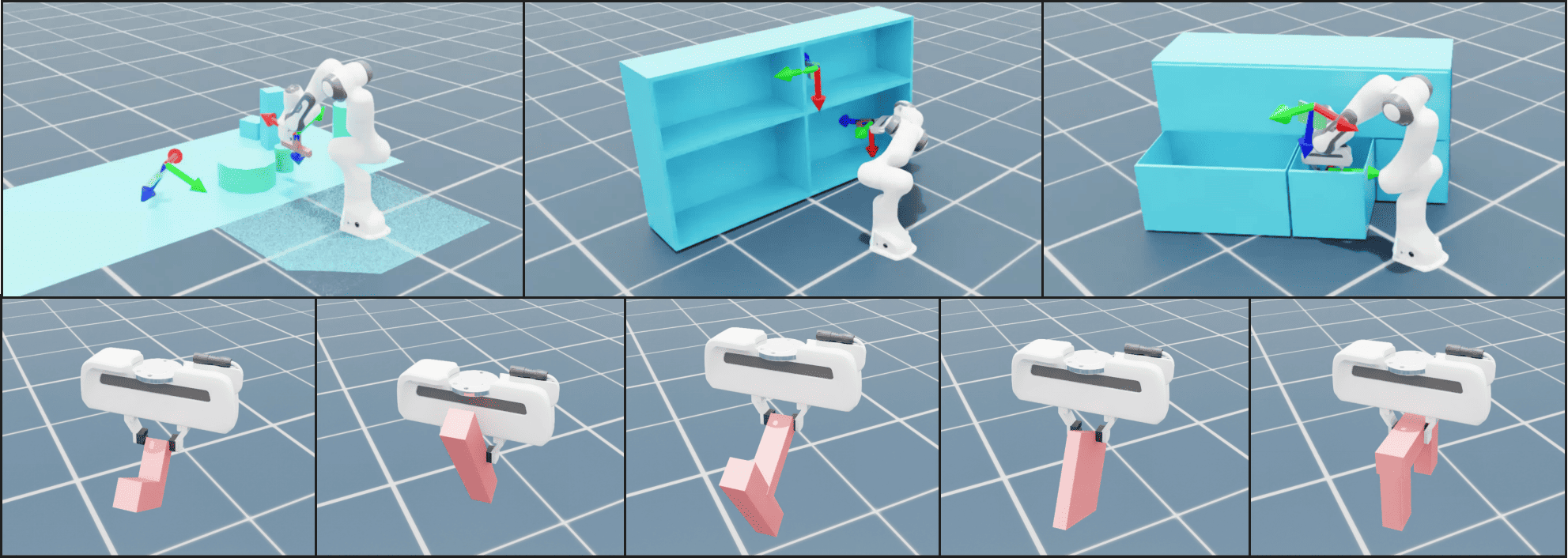}}
  \caption{\textbf{Simulated Training Environments and Tools. }
  \textbf{Top:} We utilize procedural generation to create diverse planning scenarios including (Left) Tabletop, (Center) Shelf/Cubby reaching, and (Right) Dresser collision-free goal-reaching.
  \textbf{Bottom:} Examples of procedural tool geometries}
  \label{fig:env_tool} \vspace{-5mm}
\end{figure*}

\subsection{Analytical Policy Gradient Fine-tuning}

Learning from demonstrations alone is often insufficient to ensure robust closed-loop performance, especially under distribution shift. To address covariate shift and residual collisions that remain after BC pre-training, we fine-tune the policy using Analytical Policy Gradient (APG). Specifically, we cast the kinematic motion planning process as a fully differentiable kinematic rollout (Stage 2 of Fig.~\ref{fig:method_overview}) and calculate exact gradients with respect to policy parameters $\theta$ via Backpropagation Through Time (BPTT) \cite{werbos2002backpropagation}. By avoiding expensive physics simulation and operating purely in kinematics, this procedure enables fast rollouts and efficient gradient computation.

\textbf{Fine-tuning Objectives:} During APG fine-tuning, the policy is optimized with a composite objective defined over the generated trajectory $\tau = \{\mathbf{q}_0, \dots, \mathbf{q}_{H-1}\}$, where $w_{\text{target}}$, $w_{\text{col}}$, and $w_{\text{smooth}}$ are scalar weights balancing the three terms:
\begin{equation}
\begin{aligned}
\mathcal{L}_{APG} ={}& w_{\text{target}} \mathcal{L}_{\text{target}} + w_{\text{col}} \mathcal{L}_{\text{col}} + w_{\text{smooth}} \mathcal{L}_{\text{smooth}} .
\end{aligned}
\end{equation}
The \textit{target loss} penalizes both the translational and rotational error of the end-effector pose. Here, $\mathbf{p}_{\text{pred}}, \mathbf{p}_{\text{target}} \in \mathbb{R}^3$ are the predicted and target end-effector positions, $\mathbf{R}_{\text{pred}}, \mathbf{R}_{\text{target}} \in SO(3)$ are the corresponding rotation matrices, and $\lambda_{\text{rot}}$ weights the rotational term. For the rotational term, we use the chordal distance~\cite{peretroukhin2020learning}, which yields a smooth and strictly convex error surface and avoids the vanishing-gradient issues commonly observed with quaternion-based losses~\cite{geist2024learning}:
\begin{equation}
\begin{aligned}
\mathcal{L}_{\text{target}} ={}& \left\lVert \mathbf{p}_{\text{pred}} - \mathbf{p}_{\text{target}} \right\rVert^2 + \lambda_{\text{rot}} \left\lVert \mathbf{R}_{\text{pred}} - \mathbf{R}_{\text{target}} \right\rVert_F^2.\hspace{-2mm}
\end{aligned}
\end{equation}
The \textit{collision loss} ($\mathcal{L}_{col} = \mathcal{L}_{env} + \mathcal{L}_{self}$) applies dense gradients throughout the kinematic chain. We model robot and tool geometries via approximation spheres \cite{sundaralingam2023curobo}, evaluating penetrations against the environment's differentiable Signed Distance Field (SDF), $\Phi_{env}(\mathbf{x})$, where $\mathcal{K} = \{1, \dots, K\}$ indexes the collision spheres covering all robot and tool links, $\mathbf{x}_k(\mathbf{q}_t) = \mathrm{FK}_k(\mathbf{q}_t)$ is the world-frame center of the $k$-th sphere at configuration $\mathbf{q}_t$, and $\epsilon_{\text{margin}}$ is a safety margin:
\begin{equation}
    \mathcal{L}_{env} = \sum_{t=0}^{H} \sum_{k \in \mathcal{K}} \text{ReLU}\left( \epsilon_{margin} - \Phi_{env}(\mathbf{x}_k(\mathbf{q}_t)) \right)
\end{equation}
Self-collisions ($\mathcal{L}_{self}$) are similarly penalized using a hinge loss on the signed distances calculated from SDF $\Phi_{self}(\mathbf{x})$ between all non-adjacent robot links. Specifically, for each non-adjacent link pair $(i,j)$, we compute the pairwise sphere distances and apply the same hinge formulation as Eq.~(4).

Finally, the \textit{smoothness loss} regularizes joint velocities and accelerations to ensure mechanical feasibility, where $\lambda_{\text{vel}}$ and $\lambda_{\text{acc}}$ weight the velocity and acceleration penalties:
\begin{equation}
    \mathcal{L}_{smooth} = \frac{1}{H} \sum_{t=0}^{H-1} \left( \lambda_{vel} || \Delta \mathbf{q}_t ||^2 + \lambda_{acc} || \Delta \mathbf{q}_{t+1} - \Delta \mathbf{q}_t ||^2 \right)
\end{equation}

By keeping the entire rollout fully differentiable, BPTT optimization effectively "pushes" the arm out of obstacles via the SDF gradients while "pulling" the end-effector to the target pose. The complete self-supervised APG fine-tuning is summarized in Algorithm \ref{alg:apg_finetuning}.

\subsection{Differentiable Scene Model}
The APG fine-tuning process relies on analytical gradients derived from the environment geometry. To enable this, we explicitly model all obstacles, tools, and the robot's collision geometry using primitive shapes (boxes, spheres, cylinders) within an auto-differentiation framework.

\textbf{Analytical SDF.} The environment SDF is defined as $\Phi_{\text{env}}(\mathbf{x}) = \min_{j} \phi_j(\mathbf{x})$, where $\phi_j$ is the signed distance to the $j$-th obstacle primitive. For a box with half-extents $\mathbf{h} \in \mathbb{R}^3$ centered at the origin (after applying the inverse obstacle transform), $\phi_{\text{box}}(\mathbf{x}) = \lVert \max(|\mathbf{x}| - \mathbf{h}, \mathbf{0}) \rVert + \min(\max_i(|x_i| - h_i), 0)$. For a cylinder with radius $r$ and half-height $h$, $\phi_{\text{cyl}}(\mathbf{x}) = \min\bigl(\max(\sqrt{x_1^2+x_2^2}-r,\, |x_3|-h),\, 0\bigr) + \lVert \max((\sqrt{x_1^2+x_2^2}-r,\, |x_3|-h),\, \mathbf{0}) \rVert$. Since all primitives are composed of smooth operations, the spatial gradient $\nabla_{\mathbf{x}} \Phi_{\text{env}}$ is analytically available and flows through the auto-differentiation graph.

{   \setlength{\intextsep}{0pt}
\begin{algorithm}[h]
    {
    \footnotesize
    \caption{Analytical Policy Gradient Fine-Tuning}
    \label{alg:apg_finetuning}
    \begin{algorithmic}[1]
        \State \textbf{Require:} Pre-trained Policy $\pi_\theta$, Differentiable Kinematics $\mathcal{S}$, Dataset $\mathcal{D}$
        \State \textbf{Require:} Horizon $H$, Learning Rate $\alpha$
        \State Initialize optimizer with $\alpha$, enable gradients for $\pi_\theta$

        \For{epoch $e = 1 \dots N$}
            \State Sample batch $B = \{(\mathbf{q}_{0}^{(i)}, \mathbf{T}_{ee}^{(i)}, \mathcal{O}_{env}^{(i)}) \}_{i=1}^{M}$
            \State Initialize trajectories $\mathcal{T} \leftarrow [\mathbf{q}_{0}]$

            \State \textit{// Differentiable Rollout}
            \For{$t = 0 \dots H-1$}
                \State $obs_t \leftarrow (\text{PointNet}(\mathcal{O}_{env}, \mathbf{q}_t), \mathbf{q}_t, \mathbf{T}_{ee})$
                \State $\Delta \mathbf{q}_t \leftarrow \pi_\theta(obs_t)$
                \State $\mathbf{q}_{t+1} \leftarrow \text{clamp}(\mathbf{q}_t + \Delta \mathbf{q}_t)$ \Comment{Integrate}
                \State $\mathcal{T}. \text{append}(\mathbf{q}_{t+1})$
            \EndFor

            \State \textit{// Loss Computation (BPTT)}
            \State $\mathbf{p}_{H}, \mathbf{R}_{H} \leftarrow \text{FK}(\mathbf{q}_{H})$
            \State $\mathcal{L}_{target} \leftarrow ||\mathbf{p}_{H} - \mathbf{p}_{tgt}||^2 + \lambda_{rot} ||\mathbf{R}_{H} - \mathbf{R}_{tgt}||_F^2$
            
            \State \textit{// Differentiable Collision Check}
            \State $\mathcal{L}_{col} \leftarrow \sum_{t=0}^{H} \sum_{k \in \mathcal{K}} \text{ReLU}(\epsilon_{margin} - \Phi_{env}(\mathbf{x}_{k,t}))$
            
            \State $\mathcal{L}_{smooth} \leftarrow \text{Eq. } (5)$
            \State $\mathcal{L}_{APG} \leftarrow w_{target}\mathcal{L}_{target} + w_{col}\mathcal{L}_{col} + w_{smooth}\mathcal{L}_{smooth}$
            
            \State $\theta \leftarrow \theta - \alpha \cdot \text{Clip}(\nabla_\theta \mathcal{L}_{APG})$
        \EndFor
    \end{algorithmic}
    }
\end{algorithm}}
\textbf{BPTT Gradient Path.} During APG fine-tuning, the policy is unrolled for $H$ steps: $\mathbf{q}_{t+1} = \mathbf{q}_t + \pi_\theta(\mathbf{q}_t, \mathbf{P}_t, \mathbf{T}_{ee})$. Gradients $\nabla_\theta \mathcal{L}_{\text{APG}}$ are obtained via BPTT: $\frac{\partial \mathcal{L}}{\partial \theta} = \sum_{t} \frac{\partial \mathcal{L}}{\partial \mathbf{q}_t} \frac{\partial \mathbf{q}_t}{\partial \theta}$, where the collision loss contributes via $\nabla_{\mathbf{x}} \Phi_{\text{env}}$ composed with the forward kinematics Jacobian $\frac{\partial \mathbf{x}_k}{\partial \mathbf{q}_t}$, enabling SDF gradients to ``push'' collision spheres away from obstacles.

To encourage the policy to learn robust geometric features, we further employ procedural generation to randomize workspace configurations, obstacle shapes, tool geometries, and their poses during data collection.

\section{Experiments}
We evaluate our method to answer the following questions: (i) To what extent does APG fine-tuning improve performance beyond a simple Behavior Cloning baseline? (ii) What is the benefit of explicitly representing tool geometry with point clouds relative to coarse bounding-box approximations? (iii) How does the fine-tuned policy compare with state-of-the-art classical and neural planners in terms of success rate and inference speed? (iv) Can the method adapt to unseen environments without requiring additional expert demonstrations?

\begin{table*}[t]
    \centering
    \caption{\textbf{\textsc{Policy Performance Comparison:}} We evaluate the impact of analytical policy gradient (APG) fine-tuning against a behavior cloning (BC) baseline on our and M$\pi$Net's policies. Performance is measured across 1800 held-out problems in three environments. Following the evaluation protocol of M$\pi$Nets~\cite{fishman2023motion}, we report the collision rate with the environment (Env.) and the robot itself (Self), the percentage of smooth trajectories (defined by a Spectral Arc Length~\cite{balasubramanian2015analysis} value below $-1.6$), along with position (Pos.) and orientation (Ori.) accuracy within specified error bounds. Furthermore, we evaluate the generated path length measured by total Cartesian end-effector translation (EE Pos.) and total configuration-space joint displacement (Joint). The results demonstrate that APG fine-tuning significantly improves precision and collision avoidance, and enables flexible trajectory shaping via targeted cost functions. ELMP-BC: behavior cloning only; ELMP-APG: BC + APG fine-tuning (our full method); ELMP-APG-Cartesian: ELMP-APG with an additional Cartesian path-length penalty. All variants share the same architecture and input representation; the tool representation is held fixed (point cloud) across all rows.}
    \label{tab:performance_gains}
    \resizebox{\textwidth}{!}{%
    \setlength{\tabcolsep}{5pt}
    \begin{tabular}{l||cc|c||cc||ccc||cc}
        \toprule
        \multirow{3}{*}{\textbf{Method}}
            & \multicolumn{2}{c|}{\textbf{Collision Rate} (\%) $\downarrow$}
            & \multirow{3}{*}{\textbf{Smooth} (\%) $\uparrow$}
            & \multicolumn{2}{c||}{\textbf{Pos. Accuracy} (\% within) $\uparrow$}
            & \multicolumn{3}{c||}{\textbf{Ori. Accuracy} (\% within) $\uparrow$}
            & \multicolumn{2}{c}{\textbf{Path Length} $\downarrow$} \\
        \cmidrule(lr){2-3} \cmidrule(lr){5-6} \cmidrule(lr){7-9} \cmidrule(lr){10-11}
            & \textbf{Environment} & \textbf{Self}
            &  & \textbf{1 cm} & \textbf{5 cm}
            & \textbf{5$^{\circ}$} & \textbf{15$^{\circ}$} & \textbf{30$^{\circ}$}
            & \textbf{EE Pos.} (m) & \textbf{Joint} (rad)\\
        \midrule

        \rowcolor{blue!5}
        M$\pi$Nets-BC & 16.3 & 0.6 & 95.1 & 57.6 & 83.9 & 12.8 & 57.8 & 71.2 & 0.73 & $4.30$ \\
        M$\pi$Nets-APG & \underline{10.8} & 0.4 & \underline{98.4} & 74.2 & 91.2 & 32.5 & 76.8 & 87.9 & $0.97$ & $4.31$ \\
        \midrule
        \rowcolor{blue!5}
        ELMP-BC & 31.3 & 0.7 & 92.6 & 70.4 & 82.7 & 39.8 & 76.7 & 84.7 & \underline{0.69} & $4.09$ \\

        {ELMP-APG-Cartesian} & 10.9 & \textbf{0.2} & 97.3 & \underline{86.1} & \underline{93.1} & \textbf{84.0} & \textbf{93.7} & \underline{96.5} & \textbf{0.63} & 4.10 \\

        \rowcolor{blue!5}
        {ELMP-APG} & \textbf{8.9} & 0.4 & \textbf{99.1} & \textbf{89.0} & \textbf{94.3} & 72.6 & \underline{93.5} & \textbf{96.7} & $0.85$ & \textbf{3.90} \\

        \bottomrule
    \end{tabular}
    } \vspace{-4mm}
\end{table*}

\subsection{Experimental Setup}
We evaluate our method on a test set of 1800 held-out problems across the three environments with various tools sampled from the training distribution. A trajectory is considered successful if (a) it reaches the target within a position error of 1\,cm and (b) has an orientation error of less than 15$^{\circ}$ without any collision (robot or tool). This criterion is consistent with prior learning-based motion planning benchmarks, such as M$\pi$Nets \cite{fishman2023motion} and Neural MP \cite{dalal2024neural}, ensuring that the generated plans are sufficiently precise for post-processing or direct execution in grasping tasks. Collisions are checked at discrete waypoints after 3$\times$ interpolation (${\sim}$150 configurations per trajectory) using a mesh-based checker on the full robot and tool meshes--independent of the sphere-based SDF used during training--with a safety margin of $\epsilon_{\text{margin}} = 0.03$\,m. Tool collisions are included in the reported environment collision rate.

\subsection{Implementation Details}
\textbf{Pretraining \& Data Generation:} Both our method and the M$\pi$Net baseline are pretrained via behavior cloning on a large-scale dataset of approximately 600K expert trajectories.
These trajectories were generated using AIT*~\cite{strub2020adaptively} to solve procedurally generated planning queries across three benchmark environments: \textit{Tabletop}, \textit{Cubby}, and \textit{Dresser}.

\textbf{Architecture:} A PointNet++~\cite{qi2017pointnet++} encoder with three Set Abstraction modules produces a 1024-dimensional geometric feature. Separate MLP encoders process the 7-DoF joint configuration and the 12-dimensional target pose (position + flattened rotation matrix), each producing a 64-dim embedding. The concatenated 1152-dim vector is decoded by an MLP to produce the joint increment $\Delta\mathbf{q}_t$.

\textbf{Network Inputs and Baselines:} Point cloud observations are subsampled to fixed cardinalities: $N_r = 2048$ points for the robot and tool geometry, $N_s = 4096$ for scene obstacles, and $N_t = 128$ for the target pose ($N = 6272$ total). We benchmark against AIT*~\cite{strub2020adaptively} (sampling-based), CuRobo~\cite{sundaralingam2023curobo} (optimization-based), and M$\pi$Nets~\cite{fishman2023motion} (neural). We focus on baselines with publicly available training pipelines; recent approaches such as Neural MP~\cite{dalal2024neural} and Deep Reactive Policy~\cite{yang2025deep} are omitted from the direct comparison.

\textbf{Training Stability and Optimization:} Optimizing long-horizon trajectories via BPTT is prone to vanishing or exploding gradients~\cite{bengio1994learning, pascanu2013difficulty, metz2021gradients}.
To ensure stable convergence, we employ a multi-step stabilization strategy. First, BC initialization leverages the pre-trained weights to provide a stable starting point. Second, we apply gradient norm clipping (norm $\le 1.0$) to mitigate gradient explosion during long-horizon rollouts. We use Adam with an initial learning rate of $10^{-4}$ and exponential decay, with 16-bit mixed precision on 6$\times$RTX~3090 GPUs. BC pretraining runs for 500 epochs (batch size 16); APG fine-tuning runs for 50 epochs (batch size 48, DDP) with rollout horizon $H=69$. The collision margin is $\epsilon_{\text{margin}}=0.03$\,m.

\subsection{Effectiveness of APG Fine-tuning}
We compare our APG fine-tuned policy against a BC-only baseline using the same dataset, without collecting new expert demonstrations (Tab.~\ref{tab:performance_gains}).

\noindent\emph{Pose Accuracy \& Collision Avoidance:} APG fine-tuning reduces environmental collisions by more than 3$\times$ and sharply improves positional and orientation accuracy over the BC baseline, while ensuring nearly all trajectories are smooth.

\noindent\emph{Optimization Versatility:} ELMP-APG-Cartesian, which adds an end-effector path-length penalty to $\mathcal{L}_{APG}$, produces significantly shorter paths with a minor clearance trade-off, demonstrating that the planner can be tailored to different objectives without new demonstrations.

\noindent\emph{Architectural Gains:} APG also improves the M$\pi$Nets baseline, but the decoupled pose encoder of ELMP-APG significantly outperforms M$\pi$Nets-APG in both positional success and orientation precision. The complementary role of tool-aware conditioning is examined in Sec.~\ref{sec:tool_aware}.

\begin{figure}[t]
\centering
\includegraphics[width=\columnwidth]{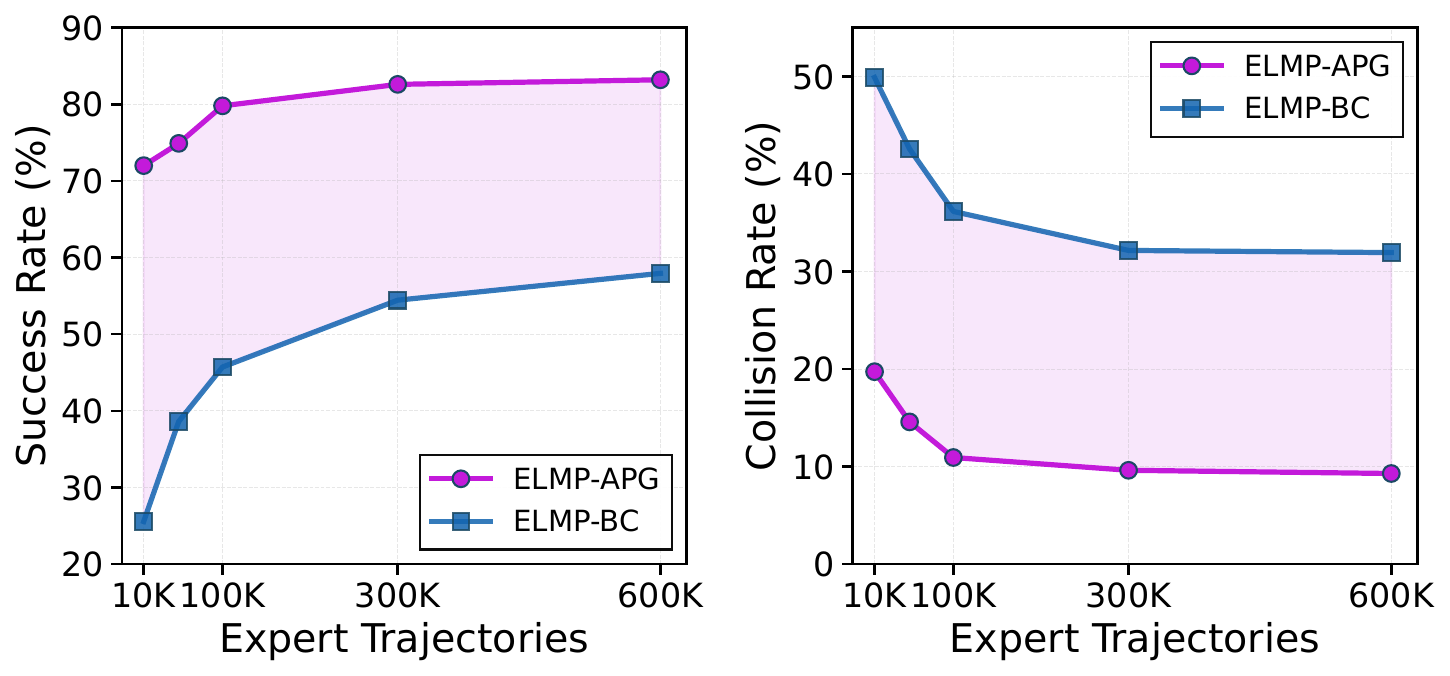}
\caption{\textbf{Pre-training data-size ablation.} Success rate (left) and environment collision rate (right) as a function of the number of expert trajectories used for BC pre-training. ELMP-APG demonstrates that APG fundamentally reduces the dependence on expert data scale.}
\label{fig:pretrain_ablation} \vspace{-7mm}
\end{figure}

\emph{Data Efficiency:} To quantify APG's dependence on expert data, we ablate the pre-training budget from 10K to 600K trajectories (Fig.~\ref{fig:pretrain_ablation}). ELMP-APG with only 10K trajectories (72.0\% success, 19.4\% collision) already surpasses ELMP-BC with the full 600K (57.9\%, 31.9\%), achieving better results with 60$\times$ less expert data. The APG gain is largest in the low-data regime.

\begin{table*}[t!]
    \centering
    \caption{\textbf{\textsc{Comparison with State-of-the-Art Planners:}} We report the success rate, average success, solution time, and cold-start (CS) time across three environments (\textit{Tabletop}, \textit{Cubby}, and \textit{Dresser}) for 1800 held-out problems. Our method is compared against AIT* (sampling-based), CuRobo (optimization-based), and M$\pi$Nets (neural). All methods are evaluated on the same 1800 tool-aware problems; this table compares across planners (architecture and input representation are fixed for each method). Our APG fine-tuned policy achieves competitive success rates while maintaining orders-of-magnitude faster solution and cold-start times.}
    \label{tab:main_results}
    \resizebox{\textwidth}{!}{%
    \setlength{\tabcolsep}{5pt}
    \begin{tabular}{l|l||ccc|c|cc}
        \toprule
        \multirow{2}{*}{\textbf{Method}}
            & \multirow{2}{*}{\textbf{Type}}
            & \multicolumn{3}{c|}{\textbf{Success Rate} (\%) $\uparrow$}
            & \multirow{2}{*}{\textbf{Average Success} (\%) $\uparrow$}
            & \multirow{2}{*}{\textbf{Sol. Time} (s) $\downarrow$}
            & \multirow{2}{*}{\textbf{CS Time} (s) $\downarrow$} \\
        \cmidrule(lr){3-5}
            &  & \textbf{Tabletop} & \textbf{Cubby} & \textbf{Dresser} &  &  &  \\
        \midrule

        AIT*~\cite{strub2020adaptively}              & Sampling      & \underline{85.3} & 64.0 & 80.8 & 76.7 & 21.5 & 21.5 \\
         \rowcolor{blue!5}
        CuRobo~\cite{sundaralingam2023curobo}        & Optimization  & 79.0 & \textbf{88.0} & \underline{91.3} & \textbf{86.1} & \textbf{0.4} & 0.4 \\

        CuRobo MPC                                   & Local         & 35.0 & 8.3 & 5.0 & 16.1 & 3.1 & 11.1\,e\text{-}3 \\
        \midrule
        \rowcolor{blue!5}
        M$\pi$Nets~\cite{fishman2023motion}           & Neural        & 32.2 & 47.0 & 71.2 & 50.1 & 0.7 & \underline{10.4\,e\text{-}3} \\
                \textbf{ELMP-APG}   & Neural        & \textbf{90.3} & \underline{72.7} & \textbf{91.5} & \underline{84.8} & \underline{0.5} & \textbf{7.8\,e\text{-}3} \\

        \bottomrule
    \end{tabular}
    }
\end{table*}

\subsection{Comparison with State-of-the-Art}

We present a quantitative comparison with the baselines in Table~\ref{tab:main_results}.
For our method and the learning-based baseline (M$\pi$Nets), we train a single unified policy on the combined dataset of all three environments.

\noindent\emph{Success Rates:} ELMP-APG achieves an 84.8\% average success rate, exceeding CuRobo in Tabletop (90.3\% vs.\ 79.0\%) and Dresser (91.5\% vs.\ 91.3\%) while remaining below it in Cubby (72.7\% vs.\ 88.0\%), and significantly surpassing AIT* (76.7\%) and M$\pi$Nets (50.1\%). This performance improvement over the M$\pi$Nets baseline highlights the direct impact of our contributions: explicit tool-aware point-cloud conditioning and self-supervised APG fine-tuning drive the massive gains in success rate. The lower baseline performance of M$\pi$Nets in our benchmarks compared to its original publication stems from a 5.5$\times$ reduction in training data scale (${\sim}$600K vs.\ 3.27M trajectories) combined with our much stricter evaluation criteria, where any collision involving the variable tool geometry results in a task failure. This indicates that ELMP-APG is more sample-efficient than M$\pi$Nets in the tool-aware setting, achieving higher success despite being trained on the reduced dataset.

\noindent\emph{Local Reactive Baseline:} CuRobo MPC, which operates as a local reactive controller, achieves only 16.1\% average success. Its myopic, single-step optimization lacks the long-horizon foresight required to navigate constrained geometries such as shelves and drawers, confirming that reactive methods alone are insufficient for these tasks.

\noindent\emph{Inference Speed and Cold Start Time:} A critical advantage of our approach is the Cold Start Time--the latency between receiving a new query and sending the first action.
Traditional global planners like AIT* and CuRobo must compute the full trajectory before execution, resulting in substantial latencies (21.5\,s and 0.4\,s, respectively).
In contrast, our neural policy achieves a cold start time of just $7.8$\,ms.
This sub-10\,ms reactivity makes our approach well suited for tasks requiring global guidance and reactive replanning in changing scenes.

\begin{table}[b!]
    \centering
\caption{\textbf{\textsc{Effectiveness of Tool-Aware Representation:}} We compare collision rates of our point-cloud encoding against a bounding-box baseline on standard and hard problem sets. Explicit point-cloud encoding better captures fine-grained geometry and reduces collisions.}
    \label{tab:tool_encoding_comparison}
    \resizebox{\columnwidth}{!}{%
    \setlength{\tabcolsep}{5pt}
    \begin{tabular}{l||cc}
        \toprule
        \multirow{2}{*}{\textbf{Method}}
            & \multicolumn{2}{c}{\textbf{Collision Rate} (\%) $\downarrow$} \\
        \cmidrule(lr){2-3}
            & \textbf{Standard Problems} & \textbf{Hard Problems} \\
        \midrule

        \rowcolor{blue!5}
        Bounding Box Feature~\cite{lee2025learning} & \underline{6.0} & \underline{30.5} \\
        \textbf{ELMP-APG} & \textbf{2.5} & \textbf{11.5} \\

        \bottomrule
    \end{tabular}
    }
\end{table}
\subsection{Effectiveness of Tool-Aware Representation}\label{sec:tool_aware}
To evaluate the effectiveness of explicitly representing tool geometry with point clouds, we compare our method with the bounding-box (BBox) abstraction employed in~\cite{lee2025learning}. We consider two evaluation settings. The \emph{Standard Problem} is drawn from the same distribution as the data generation process and captures nominal operating conditions.
The \emph{Hard Problem} consists of adversarially sampled target poses where we alter the safety margin and distance thresholds to sample the end-effector and tool poses within a $-3$\,cm to 10\,cm range from environmental obstacles. The negative bound allows minor penetration between the target and the obstacles, requiring more precise collision reasoning.

The rationale for designing the \textit{Hard Problem} set is to determine if the policy truly considers the tool geometry. Even if a policy ignores tool geometry, it can still learn the general trajectory distribution and succeed in nominal cases. However, in the \textit{Hard} scenario, a policy that relies on distribution memorization rather than geometric awareness will fail to adjust for the specific tool-obstacle conflict.

As shown in Table \ref{tab:tool_encoding_comparison}, our point cloud embedding significantly outperforms the bounding box baseline.
In the \textit{Standard} problem, the collision rate is reduced from 6.0\% to 2.5\%.
This gap widens significantly on the \textit{Hard} problem, where the BBox baseline suffers a collision rate of 30.5\% compared to 11.5\% for our method.
This suggests that explicit point cloud encoding enables the policy to resolve fine-grained geometric conflicts that coarse bounding box abstractions fail to capture.

\subsection{Self-Supervised Transfer to Novel Environments}

\begin{figure}[tb]\centering\includegraphics[width=0.47\textwidth]{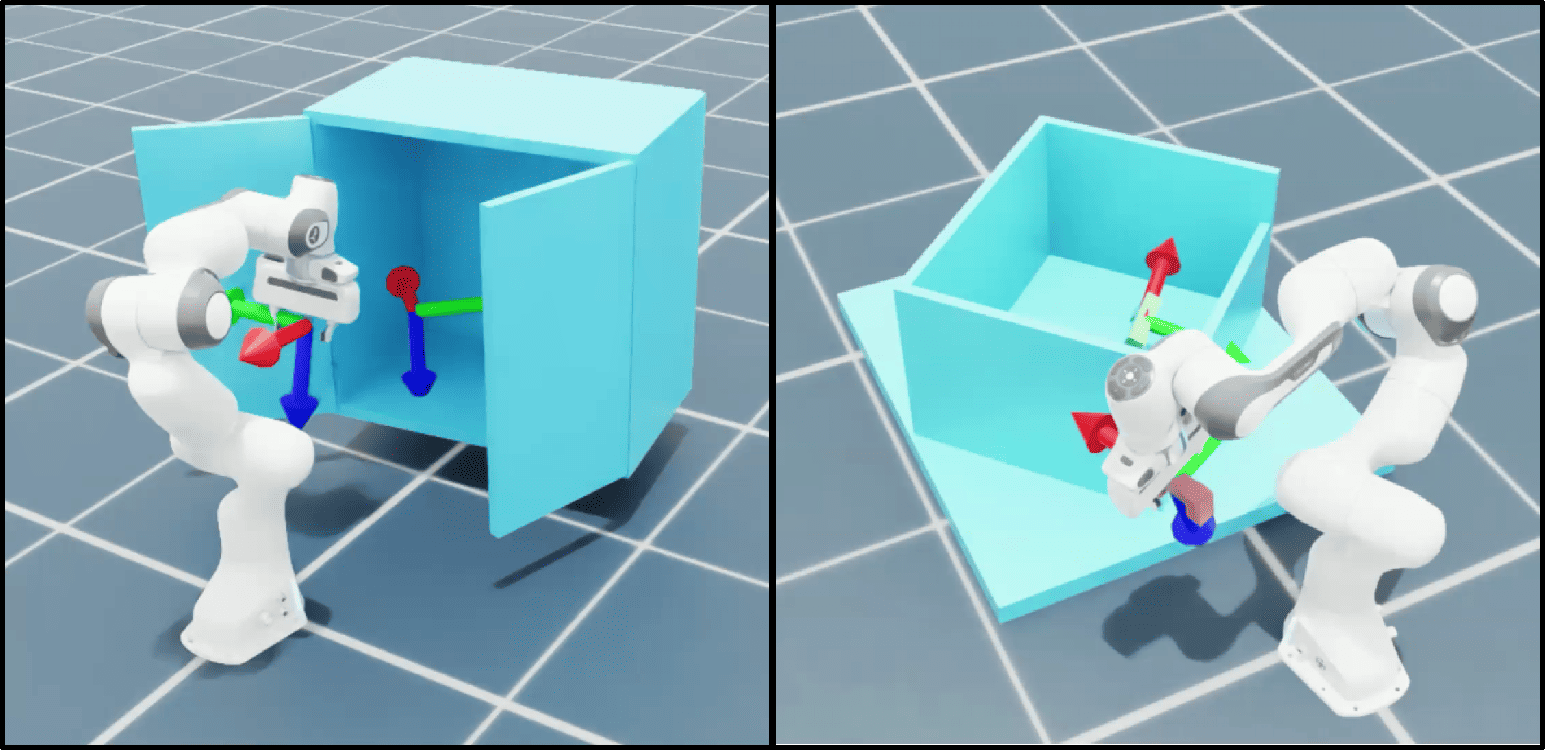}
\caption{\textbf{Novel unseen environments} for transfer evaluation: \textit{Cabinet} (left) and \textit{Bin} (right), with geometric structures not observed during pre-training.}\label{fig:APG}\end{figure}

We evaluate transfer to two novel environments--\textit{Cabinet} and \textit{Bin} (Fig.~\ref{fig:APG})--by sampling random start-goal pairs (checked for IK feasibility) and fine-tuning with APG, without collecting new expert demonstrations. Procedural problem sampling takes only 82\,ms per scenario, reducing per-sample cost by two orders of magnitude versus AIT* (21.5\,s).

\begin{figure}[tb]
\centering
\includegraphics[width=1.0\linewidth]{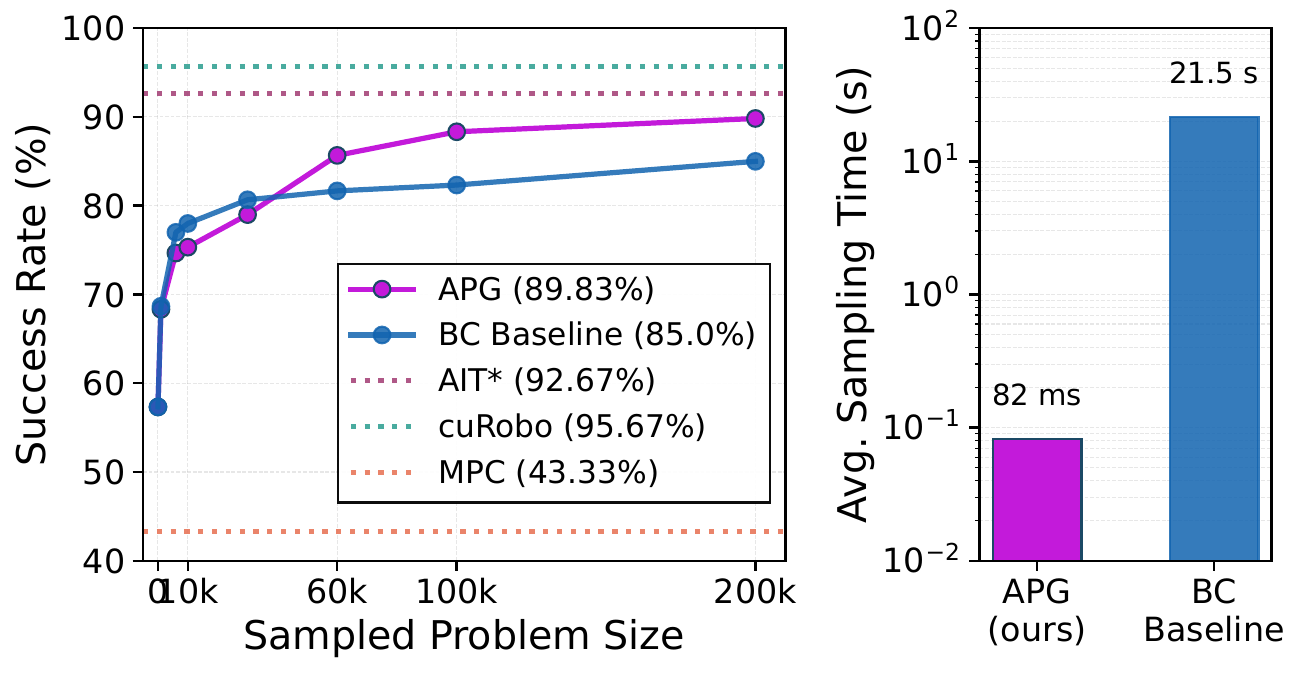}
\caption{\textbf{Fine-tuning performance in novel environments.}
APG steadily increases success rates in the unseen Cabinet and Bin environments as the number of sampled self-supervised problems increases.
While APG and BC share similar learning curves, APG ultimately plateaus at a higher success rate (89.8\% vs 85.0\%) and requires two orders of magnitude less data generation time.}
\label{fig:adaptation_results}
\end{figure}

As shown in Fig.~\ref{fig:adaptation_results}, APG converges to a higher success rate than a BC baseline trained on expert trajectories for the same problems (89.8\% vs.\ 85.0\%), starting from a 57.3\% zero-shot baseline. The computational advantage is stark: sampling 200K problems for APG takes 4.5 CPU-hours (under 3 minutes), while generating the same number of expert trajectories for BC demands over 1,194 CPU-hours.

\subsection{Real-World Deployment}

We validated our policy on a physical Franka Emika Panda with a fixed RealSense depth camera.
The analytic SDF is used only during APG fine-tuning; at deployment, the policy consumes only a point cloud and target pose.
We filter robot/tool geometry from the raw depth image using \texttt{realtime\_urdf\_filter}~\cite{blodow_realtime_urdf_filter_2022}, then aggregate remaining scene points with internally generated robot and tool points via proprioception, ensuring precise tool-geometry awareness despite occlusion.

\begin{figure}[tb]
    \centering
    \includegraphics[width=0.95\linewidth]{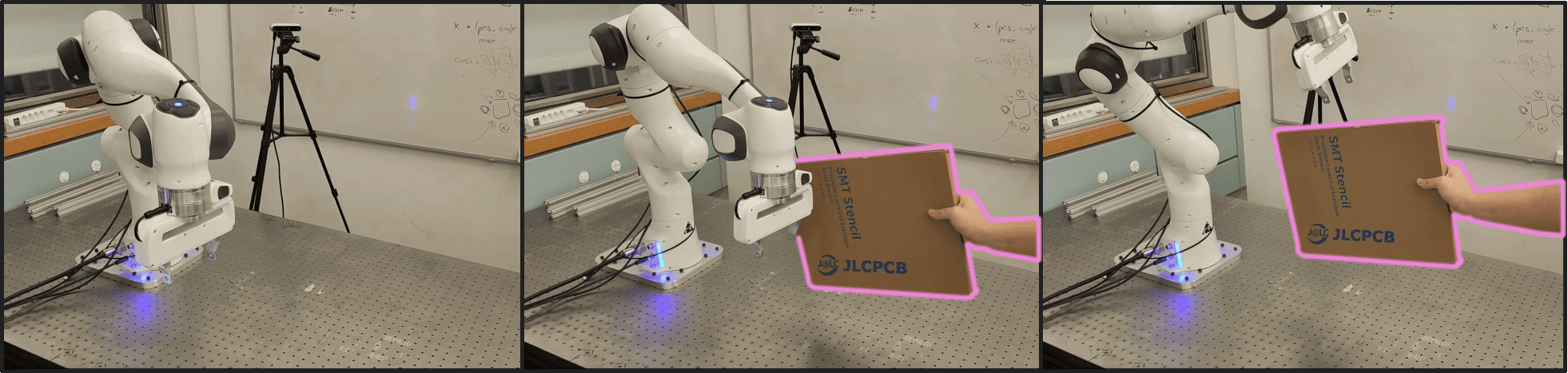} \\
    \vspace{1pt}
    {\footnotesize (a) \textbf{Reactive Replanning:} Reactive replanning around moving obstacles (highlighted in pink). } \\
    \vspace{3pt}
    \includegraphics[width=0.95\linewidth]{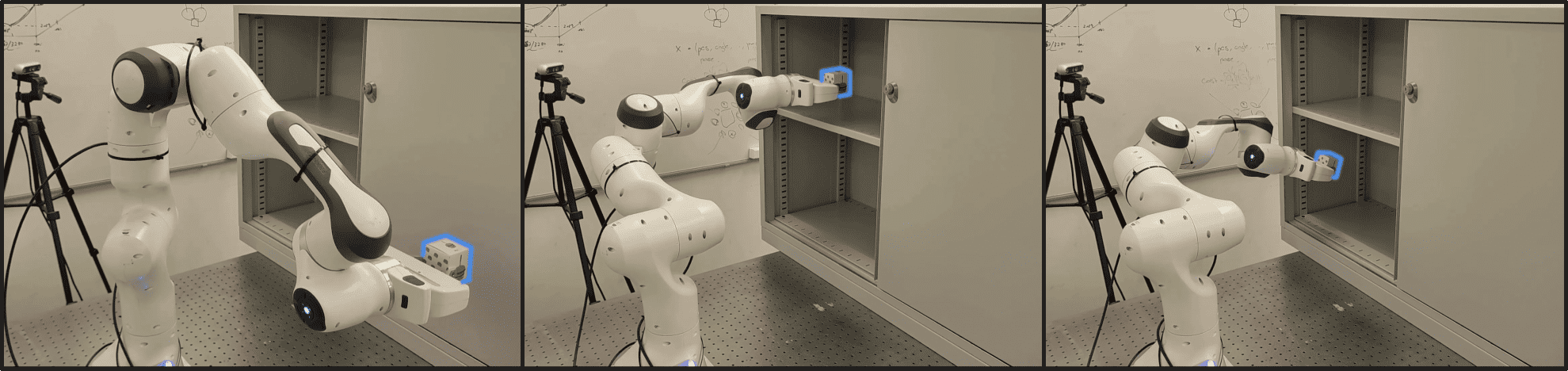} \\
    \vspace{1pt}
    {\footnotesize (b) \textbf{In-Distribution:} Goal reaching with a shelf environment. } \\
    \vspace{3pt}
    \includegraphics[width=0.95\linewidth]{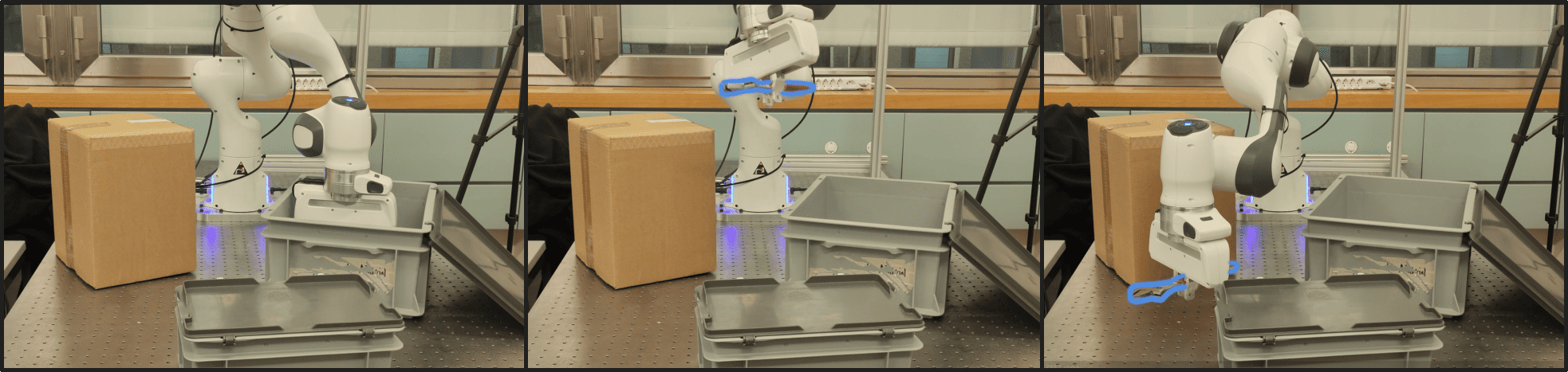} \\
    \vspace{1pt}
    {\footnotesize (c) \textbf{Unseen Environment Transfer:} Collision-free placement in a novel bin task. }

    \caption{Real-world deployment of the APG fine-tuned policy on a Franka Emika Panda robot. The moving obstacles are highlighted in pink (a), and the tool geometry is outlined in blue (b, c). }
    \label{fig:real_world_combined}
\end{figure}

Fig.~\ref{fig:real_world_combined} shows reactive replanning around moving obstacles~(a), in-distribution cubby reaching~(b), and transfer to a novel unseen bin~(c). The policy runs in an MPC-style closed loop at 30\,Hz with 10\,Hz perception updates, treating each point cloud as a static snapshot.
To quantify performance, we conducted 36 trials across three scenes and three tools (Table~\ref{tab:real_world}), with three waypoints per step linearly interpolated and sent to the Franka's position controller.

\begin{table}[b]
    \centering\vspace{-4mm}
\caption{\textbf{\textsc{Real-World Hardware Evaluation:}} We evaluate ELMP-APG on 36 physical-robot trials across three deployment scenes (Shelf, Clutter, Bin) and three tool configurations. Success and collision rates are reported over all trials, while final position and orientation errors are averaged over successful executions only.}
    \label{tab:real_world}
    \resizebox{\columnwidth}{!}{%
    \setlength{\tabcolsep}{5pt}
    \begin{tabular}{l||c|c|c|c}
        \toprule
        \textbf{Scene} & \textbf{Success (\%)} $\uparrow$ & \textbf{Pos.\ Err.\ (cm)} $\downarrow$ & \textbf{Ori.\ Err.\ ($^{\circ}$)} $\downarrow$ & \textbf{Collision (\%)} $\downarrow$ \\
        \midrule
        \rowcolor{blue!5}
        Shelf       & 58.3 & 2.54 & 14.15 & 33.3 \\
        Clutter     & 83.3 & 1.52 & 7.60  & 16.7 \\
        \rowcolor{blue!5}
        Bin         & 75.0 & 1.93 & 6.28  & 25.0 \\
        \midrule
        \textbf{Overall} & \textbf{72.2} & \textbf{1.94} & \textbf{8.91} & \textbf{25.0} \\
        \bottomrule
    \end{tabular}
    }
\end{table}

The overall 72.2\% success rate with reasonable accuracy (1.94\,cm, 8.91$^{\circ}$) confirms hardware transfer. The collision rate increases from 8.9\% in simulation to 25.0\% on the real robot. This gap is primarily attributable to depth-camera noise and self-occlusion: the single fixed camera produces incomplete point clouds when the robot arm occudes parts of the scene, and depth artifacts near object edges introduce phantom geometry. These perception errors are not modeled during APG fine-tuning, which operates on clean analytic primitives. Incorporating realistic sensor noise models during training or employing temporal point-cloud fusion and Teacher-Student distillation~\cite{chen2020learning, lee2020learning} are promising directions to close this gap (Sec.~\ref{sec:conclusion}).

\section{Conclusion and Future Work}\label{sec:conclusion}
We presented \textbf{ELMP}, a tool-aware neural motion planner combining APG fine-tuning with explicit point-cloud tool conditioning for fast, robust collision avoidance, with qualitative evidence of reactive behavior in changing scenes. Self-supervised APG enables data-efficient adaptation to novel environments, reducing data generation costs by two orders of magnitude. This approach approaches the average success rate of optimization-based planners such as CuRobo, exceeding it in open environments while remaining below it in highly constrained settings, while maintaining millisecond-level inference latency.

Several avenues remain for future work: bridging the sim-to-real perception gap via temporal point-cloud fusion or Teacher-Student distillation~\cite{chen2020learning, lee2020learning}; handling non-primitive obstacles through differentiable collision penalties on raw point clouds or neural SDFs~\cite{bukhari2025differentiable}; extending beyond fixed-base manipulators to contact-heavy systems, likely requiring advanced APG algorithms such as SHAC~\cite{xu2022accelerated}; and integrating time-optimal path parameterization (e.g., TOPP-RA~\cite{pham2018new}) for hard velocity and acceleration guarantees.

\bibliographystyle{IEEEtran}
\bibliography{ref}

@article{pham2018new,
  title={A new approach to time-optimal path parameterization based on reachability analysis},
  author={Pham, Hung and Pham, Quang-Cuong},
  journal={IEEE Transactions on Robotics},
  volume={34},
  number={3},
  pages={645--659},
  year={2018},
  publisher={IEEE}
}

@article{qureshi2021motion,
  title={Motion planning networks: Bridging the gap between learning-based and classical motion planners},
  author={Qureshi, Ahmed H and Miao, Yinglong and Simeonov, Anthony and Yip, Michael C},
  journal={IEEE Transactions on Robotics},
  volume={37},
  number={1},
  pages={48--66},
  year={2021},
  publisher={IEEE}
}

@inproceedings{bukhari2025differentiable,
  title={Differentiable Composite Neural Signed Distance Fields for Navigation in Dynamic Indoor Scenes},
  author={Bukhari, Hasan Asy'ari and Tran, Minh-Quan Viet and Saputra, Muhamad Risqi U and Markham, Andrew},
  booktitle={IEEE International Conference on Robotics and Automation (ICRA)},
  year={2025},
  organization={IEEE}
}

@misc{blodow_realtime_urdf_filter_2022,
  author = {blodow},
  month = {12},
  title = {{realtime\_urdf\_filter}},
  url = {https://github.com/blodow/realtime\_urdf\_filter},
  version = {main},
  year = {2022}
}

@article{balasubramanian2015analysis,
  title={On the analysis of movement smoothness},
  author={Balasubramanian, Sivakumar and Melendez-Calderon, Alejandro and Roby-Brami, Agnes and Burdet, Etienne},
  journal={Journal of neuroengineering and rehabilitation},
  volume={12},
  number={1},
  pages={112},
  year={2015},
  publisher={Springer}
}

@inproceedings{fishman2023motion,
  title={Motion policy networks},
  author={Fishman, Adam and Murali, Adithyavairavan and Eppner, Clemens and Peele, Bryan and Boots, Byron and Fox, Dieter},
  booktitle={Conference on Robot Learning},
  pages={967--977},
  year={2023},
  organization={PMLR}
}

@article{dalal2024neural,
  title={Neural mp: A generalist neural motion planner},
  author={Dalal, Murtaza and Yang, Jiahui and Mendonca, Russell and Khaky, Youssef and Salakhutdinov, Ruslan and Pathak, Deepak},
  journal={arXiv preprint arXiv:2409.05864},
  year={2024}
}

@article{yang2025deep,
  title={Deep Reactive Policy: Learning Reactive Manipulator Motion Planning for Dynamic Environments},
  author={Yang, Jiahui and Liu, Jason Jingzhou and Li, Yulong and Khaky, Youssef and Shaw, Kenneth and Pathak, Deepak},
  journal={arXiv preprint arXiv:2509.06953},
  year={2025}
}

@article{werbos2002backpropagation,
  title={Backpropagation through time: what it does and how to do it},
  author={Werbos, Paul J},
  journal={Proceedings of the IEEE},
  volume={78},
  number={10},
  pages={1550--1560},
  year={2002},
  publisher={IEEE}
}

@article{lavalle1998rapidly,
  title={Rapidly-exploring random trees: A new tool for path planning},
  author={LaValle, Steven},
  journal={Research Report 9811},
  year={1998},
  publisher={Department of Computer Science, Iowa State University}
}

@article{kavraki1996probabilistic,
  title={Probabilistic roadmaps for path planning in high-dimensional configuration spaces},
  author={Kavraki, Lydia E and Svestka, Petr and Latombe, J-C and Overmars, Mark H},
  journal={IEEE transactions on Robotics and Automation},
  volume={12},
  number={4},
  pages={566--580},
  year={1996},
  publisher={IEEE}
}

@inproceedings{bohlin2000path,
  title={Path planning using lazy PRM},
  author={Bohlin, Robert and Kavraki, Lydia E},
  booktitle={Proceedings 2000 ICRA. Millennium conference. IEEE international conference on robotics and automation. Symposia proceedings (Cat. No. 00CH37065)},
  volume={1},
  pages={521--528},
  year={2000},
  organization={IEEE}
}

@article{gammell2020batch,
  title={Batch informed trees (bit*): Informed asymptotically optimal anytime search},
  author={Gammell, Jonathan D and Barfoot, Timothy D and Srinivasa, Siddhartha S},
  journal={The International Journal of Robotics Research},
  volume={39},
  number={5},
  pages={543--567},
  year={2020},
  publisher={SAGE Publications Sage UK: London, England}
}

@article{pan2025learning,
  title={Learning on the Fly: Rapid Policy Adaptation via Differentiable Simulation},
  author={Pan, Jiahe and Xing, Jiaxu and Reiter, Rudolf and Zhai, Yifan and Aljalbout, Elie and Scaramuzza, Davide},
  journal={arXiv preprint arXiv:2508.21065},
  year={2025}
}

@inproceedings{strub2020adaptively,
  title={Adaptively Informed Trees (AIT*): Fast asymptotically optimal path planning through adaptive heuristics},
  author={Strub, Marlin P and Gammell, Jonathan D},
  booktitle={2020 IEEE International Conference on Robotics and Automation (ICRA)},
  pages={3191--3198},
  year={2020},
  organization={IEEE}
}

@article{liu2018convex,
  title={The convex feasible set algorithm for real time optimization in motion planning},
  author={Liu, Changliu and Lin, Chung-Yen and Tomizuka, Masayoshi},
  journal={SIAM Journal on Control and optimization},
  volume={56},
  number={4},
  pages={2712--2733},
  year={2018},
  publisher={SIAM}
}

@article{schulman2014motion,
  title={Motion planning with sequential convex optimization and convex collision checking},
  author={Schulman, John and Duan, Yan and Ho, Jonathan and Lee, Alex and Awwal, Ibrahim and Bradlow, Henry and Pan, Jia and Patil, Sachin and Goldberg, Ken and Abbeel, Pieter},
  journal={The International Journal of Robotics Research},
  volume={33},
  number={9},
  pages={1251--1270},
  year={2014},
  publisher={Sage Publications Sage UK: London, England}
}

@inproceedings{ratliff2009chomp,
  title={CHOMP: Gradient optimization techniques for efficient motion planning},
  author={Ratliff, Nathan and Zucker, Matt and Bagnell, J Andrew and Srinivasa, Siddhartha},
  booktitle={2009 IEEE International Conference on Robotics and Automation},
  pages={489--494},
  year={2009},
  organization={IEEE}
}

@inproceedings{sundaralingam2023curobo,
  title={Curobo: Parallelized collision-free robot motion generation},
  author={Sundaralingam, Balakumar and Hari, Siva Kumar Sastry and Fishman, Adam and Garrett, Caelan and Van Wyk, Karl and Blukis, Valts and Millane, Alexander and Oleynikova, Helen and Handa, Ankur and Ramos, Fabio and others},
  booktitle={2023 IEEE International Conference on Robotics and Automation (ICRA)},
  pages={8112--8119},
  year={2023},
  organization={IEEE}
}

@article{freeman2021brax,
  title={Brax-a differentiable physics engine for large scale rigid body simulation, 2021},
  author={Freeman, C Daniel and Frey, Erik and Raichuk, Anton and Girgin, Sertan and Mordatch, Igor and Bachem, Olivier},
  journal={URL http://github. com/google/brax},
  volume={6},
  year={2021}
}

@misc{newton,
  title = {{Newton}: {GPU}-accelerated physics simulation for robotics, and simulation research.},
  author = {{Newton Contributors}},
  year = {2025},
  url = {https://github.com/newton-physics/newton},
  organization = {{Newton a Series of LF Projects, LLC}},
  license = {Apache-2.0}
}

@inproceedings{xu2022accelerated,
  title={Accelerated Policy Learning with Parallel Differentiable Simulation},
  author={Xu, Jie and Makoviychuk, Viktor and Narang, Yashraj and Ramos, Fabio and Matusik, Wojciech and Garg, Animesh and Macklin, Miles},
  booktitle={International Conference on Learning Representations (ICLR)},
  year={2022}
}

@inproceedings{Wiedemann2023Training,
    title = {Training Efficient Controllers via Analytic Policy Gradient},
    author = {Wiedemann, Nina and W{\"u}est, Valentin and Loquercio, Antonio and M{\"u}ller, Matthias and Floreano, Dario and Scaramuzza, Davide},
    booktitle = {2023 IEEE International Conference on Robotics and Automation (ICRA)},
    year = {2023},
    pages = {1349--1356},
    doi = {10.1109/ICRA48891.2023.10160581}
}

@article{lee2025learning,
  title={Learning Fast, Tool-Aware Collision Avoidance for Collaborative Robots},
  author={Lee, Joonho and Kim, Yunho and Kim, Seokjoon and Nguyen, Quan and Heo, Youngjin},
  journal={IEEE Robotics and Automation Letters},
  year={2025},
  publisher={IEEE}
}

@article{bengio1994learning,
  title={Learning long-term dependencies with gradient descent is difficult},
  author={Bengio, Yoshua and Simard, Patrice and Frasconi, Paolo},
  journal={IEEE transactions on neural networks},
  volume={5},
  number={2},
  pages={157--166},
  year={1994},
  publisher={IEEE}
}

@article{metz2021gradients,
  title={Gradients are not all you need},
  author={Metz, Luke and Freeman, C Daniel and Schoenholz, Samuel S and Kachman, Tal},
  journal={arXiv preprint arXiv:2111.05803},
  year={2021}
}

@inproceedings{bhardwaj2022storm,
  title={Storm: An integrated framework for fast joint-space model-predictive control for reactive manipulation},
  author={Bhardwaj, Mohak and Sundaralingam, Balakumar and Mousavian, Arsalan and Ratliff, Nathan D and Fox, Dieter and Ramos, Fabio and Boots, Byron},
  booktitle={Conference on Robot Learning},
  pages={750--759},
  year={2022},
  organization={PMLR}
}

@article{van2022geometric,
  title={Geometric fabrics: Generalizing classical mechanics to capture the physics of behavior},
  author={Van Wyk, Karl and Xie, Mandy and Li, Anqi and Rana, Muhammad Asif and Babich, Buck and Peele, Bryan and Wan, Qian and Akinola, Iretiayo and Sundaralingam, Balakumar and Fox, Dieter and others},
  journal={IEEE Robotics and Automation Letters},
  volume={7},
  number={2},
  pages={3202--3209},
  year={2022},
  publisher={IEEE}
}

@article{qi2017pointnet++,
  title={Pointnet++: Deep hierarchical feature learning on point sets in a metric space},
  author={Qi, Charles Ruizhongtai and Yi, Li and Su, Hao and Guibas, Leonidas J},
  journal={Advances in neural information processing systems},
  volume={30},
  year={2017}
}

@article{peretroukhin2020learning,
  title={Learning with 3D Rotations: A Hitchhiker's Guide to SO(3)},
  author={Peretroukhin, Valentin and Giamou, Matthew and Rosen, David M and Greene, W Nicholas and Roy, Nicholas and Kelly, Jonathan},
  journal={arXiv preprint arXiv:2006.01031},
  year={2020}
}

@inproceedings{pascanu2013difficulty,
  title={On the difficulty of training recurrent neural networks},
  author={Pascanu, Razvan and Mikolov, Tomas and Bengio, Yoshua},
  booktitle={International Conference on Machine Learning (ICML)},
  pages={1310--1318},
  year={2013}
}

@article{geist2024learning,
  title={Learning with 3d rotations, a hitchhiker's guide to so (3)},
  author={Geist, A Ren{\'e} and Frey, Jonas and Zhobro, Mikel and Levina, Anna and Martius, Georg},
  journal={arXiv preprint arXiv:2404.11735},
  year={2024}
}

@inproceedings{chen2020learning,
  title={Learning by Cheating},
  author={Chen, Dian and Zhou, Brady and Koltun, Vladlen and Kr{\"a}henb{\"u}hl, Philipp},
  booktitle={Conference on Robot Learning},
  pages={66--75},
  year={2020},
  organization={PMLR}
}

@article{lee2020learning,
  title={Learning quadrupedal locomotion over challenging terrain},
  author={Lee, Joonho and Hwangbo, Jemin and Wellhausen, Lorenz and Koltun, Vladlen and Hutter, Marco},
  journal={Science robotics},
  volume={5},
  number={47},
  pages={eabc5986},
  year={2020},
  publisher={American Association for the Advancement of Science}
}

\end{document}